%% file: main.tex
\documentclass[letterpaper, 10 pt, conference, article]{ieeeconf}

\IEEEoverridecommandlockouts 

\usepackage{graphicx}
\usepackage{amsmath}
\usepackage{amssymb}
\usepackage{booktabs}  
\usepackage{float}     
\usepackage{array} 
\usepackage{censor}
\usepackage{hyperref}

\usepackage[table]{xcolor} 

\title{\LARGE \bf $M^2$-VLA: Boosting Vision-Language Models for Generalizable Manipulation via Layer Mixture and Meta-Skills
}
\author{{Siyao Xiao$^{1,2}$, Yuhong Zhang$^{1}$, Zhifang Liu$^{1}$, Zihan Gao$^{1}$, Jingye Zhang$^{1}$, Sinwai Choo$^{1}$, Dake Zhong$^{1}$,}  \\
{Mengzhe Wang$^{3}$, Xiao Lin$^{4}$, Xianfeng Zhou$^{1}$, Jia Jia$^{2*}$, and Haoqian Wang$^{1*}$}
\thanks{*{Corresponding author.(email: wanghaoqian@sz.tsinghua.edu.cn; jiaj@pcl.ac.cn)}}
\thanks{$^{1}${Tsinghua Shenzhen International Graduate School, Tsinghua University, Shenzhen, China.}}%
\thanks{$^{2}${PengCheng Laboratory, Shenzhen, China.}}%
\thanks{$^{3}${College of Intelligent Robotics and Advanced Manufacturing, Fudan University, Shanghai, China.}}%
\thanks{$^{4}${Synapath.}}%
\thanks{This work was supported by the National Natural Science Foundation of China under Grants (62576190), in part by the Shenzhen Science and Technology Project under Grant KJZD20240903103210014, and in part by The Major Key Project of PCL (PCL2025A17-4).}%
}

\begin{document}
        \maketitle
        \thispagestyle{empty}
        \pagestyle{empty}

        \begin{abstract}
                Current Vision-Language-Action (VLA) models predominantly rely on end-to-end fine-tuning. While effective, this paradigm compromises the inherent generalization capabilities of Vision-Language Models (VLMs) and incurs catastrophic forgetting. To address these limitations, we propose $M^2$-VLA,  which demonstrates that a generalized VLM is able to serve as a powerful backbone for robotic manipulation directly. However, it remains a key challenge to bridge the gap between the high-level semantic understanding of VLMs and the precise requirements of robotic control. To overcome this, we introduce the Mixture of Layers (MoL) strategy that selectively extracts task-critical information from dense semantic features. Furthermore, to facilitate efficient trajectory learning under constrained model capacity, we propose a Meta Skill Module (MSM) that integrates strong inductive biases. Extensive experiments in both simulated and real-world environments demonstrate the effectiveness of our approach. Furthermore, generalization and ablation studies validate the architecture's zero-shot capabilities and confirm the contribution of each key component. The code is publicly available at \href{https://github.com/ShowiBin/M2-VLA}{github.com/ShowiBin/M2-VLA}.            
        \end{abstract}

        \input{sections/intro.tex}

        \input{sections/related.tex}
        \input{sections/method.tex}
        \input{sections/exp.tex}
\section{Conclusion}
        In this work, we introduce $M^2$-VLA, a novel architectural paradigm that enables generalized Vision-Language Models to serve as powerful backbones for Vision-Language-Action models. By incorporating the Mixture of Layers (MoL) mechanism to extract task-critical information and the Meta Skill Module (MSM) to facilitate efficient trajectory learning under limited model capacity, $M^2$-VLA achieves strong generalization performance on both instruction-following and novel-object manipulation tasks. In future work, we plan to equip the framework with multiple action heads tailored to different tasks and robot embodiments. By combining a shared brain with multiple specialized controller we aim to move toward a universal VLA model.

        \bibliographystyle{IEEEtran}
        \bibliography{sample}

\end{document}

%% file: sections/intro.tex
\section{INTRODUCTION}
\label{sec:intro}

Achieving generalizable robot manipulation remains a fundamental challenge in Embodied AI. Recently, the remarkable success of Vision-Language Models (VLMs) in general-purpose scene understanding \cite{liu2024llava, alayrac2022flamingo} has catalyzed the development of Vision-Language-Action (VLA) models \cite{brohan2023rt2, kim2024openvla,zhang2025robowheel}, which demonstrate impressive capabilities in language-conditioned manipulation tasks.
        
\begin{figure}[thpb]
        \centering
        \includegraphics[width=0.95\linewidth]{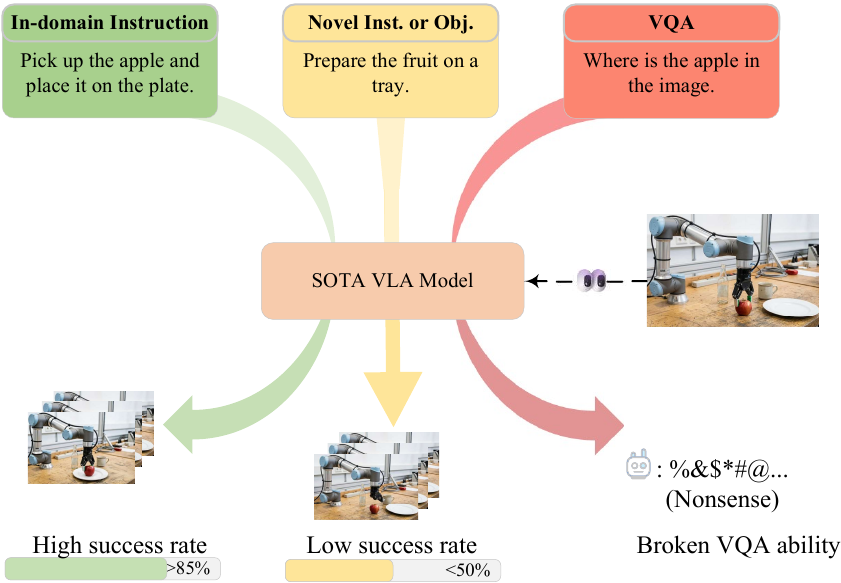}
        \caption{The catastrophic forgetting phenomenon of VLM backbone in VLA. While SOTA VLA models perform well on in-domain instructions or tasks, they struggle with novel instructions or tasks and suffer from severe degradation of VQA capabilities.}
        \label{fig:problem}
\end{figure}

Current VLA frameworks typically utilize pre-trained VLM backbones, fine-tuning their internal parameters on large-scale robotic datasets to repurpose the model's generation capabilities from language tokens to action tokens. However, this paradigm brings catastrophic forgetting \cite{luo2023empirical}, degrading the VLM's inherent semantic understanding and consequently limiting the VLA's generalizability \cite{poria202510, yu2025survey}. As illustrated in Fig.~\ref{fig:problem}, while state-of-the-art VLA models perform well on in-domain tasks, they struggle with novel instructions or objects and frequently lose their Visual Question Answering (VQA) capabilities.

We argue that preserving semantic understanding is a fundamental prerequisite for robust manipulation \cite{black2024pi0, brohan2023rt2}. Inspired by the efficient adaptation strategies for VLMs \cite{ zhang2023llamaadapter, chen2024bayesian, zeng2024enhancing}, we propose that a generalized VLM can be effectively adapted as the backbone for robotic manipulation without compromising its core capabilities. However, a fundamental gap exists between the high-level semantic reasoning of VLMs and the fine-grained control required for robotics. To bridge this, we introduce the Mixture of Layers (MoL) strategy, which aggregates information across network layers to extract task-critical features. Furthermore, retaining a generalized VLM backbone inherently restricts the number of trainable parameters, thereby limiting the overall model capacity. To enhance trajectory learning under constrained model capacity, we propose the Meta Skill Module (MSM), which leverages knowledge reuse to introduce strong inductive biases. Extensive experiments in both simulated and real-world environments validate the effectiveness of our approach, while visualization and ablation studies confirm the generalization ability and contribution of each component in overcoming performance bottlenecks.

\begin{itemize}
                
        \item We introduce the Mixture of Layers (MoL) mechanism to bridge high-level semantics and low-level control by extracting manipulation-critical fine-grained spatial information from VLM representations.

        \item We propose the Meta Skill Module (MSM), which enhances trajectory learning with a lightweight action head by promoting knowledge reuse and injecting strong inductive biases.
        
        \item The simulation and real-world experiments demonstrate that, through a novel architectural design, a generalized VLM can be adapted to serve as a backbone for high-performance robotic manipulation.


\end{itemize}

%% file: sections/related.tex
\section{RELATED WORKS}
        \subsection{Generalist Robot Policies}
        The pursuit of generalist robot policies has shifted the paradigm from task-specific engineering to large-scale imitation learning. Early approaches primarily relied on Behavior Cloning (BC) \cite{hussein2023robust} or Reinforcement Learning (RL) \cite{luo2024precise} trained on domain-specific datasets. While effective in controlled environments, these methods struggle to generalize to unseen objects or scenes. Recent advancements have demonstrated the effectiveness of Transformer-based architectures in large-scale robot learning. Notable works such as Gato \cite{gato} and RT-2 \cite{brohan2023rt2} unify multi-modal inputs (i.e., images, text, and proprioception) into a single sequence, enabling a single policy to perform diverse tasks. Furthermore, initiatives like Octo \cite{octo} and Open X-Embodiment \cite{open_x} have highlighted the potential of training across heterogeneous robot embodiments. In contrast, our work adheres to the Vision-Language-Action (VLA) paradigm, employing a Large Language Model (LLM) backbone to uniformly encode multi-modal inputs for action generation. This approach allows us to directly leverage the robust generalization capabilities inherent in pre-trained VLMs.

        \subsection{Vision-Language-Action Models}
        VLA models \cite{black2024pi0, liu2024rdt1b, zhao2025cotvla, ge2025vggtdp} are designed to integrate visual observations and proprioceptive data with natural language instructions to generate low-level robotic control signals. Capitalizing on the extensive pre-training of Vision-Language Models (VLMs) \cite{li2025eagle2, goldberg2025robo2vlm, karamcheti2024prismatic}, VLAs aim to transfer strong generalization capabilities to physical control. Several studies \cite{molmoact2025, thinkact2025} have reported VLA's strong performance on authoritative simulation benchmarks, such as LIBERO \cite{libero} and CALVIN \cite{calvin}, as well as in real-world settings. A prevailing trend involves fine-tuning the VLM backbone on large-scale robotic datasets to align representations with control tasks. Although existing literature suggests that fine-tuned VLMs adapt better to manipulation tasks, this paradigm is data-intensive and often induces catastrophic forgetting, degrading the backbone's original vision-language reasoning capabilities. Challenging this convention, we posit that  a generalized VLM can be adapted to serve
        as a robust backbone for robotic manipulation.

        \subsection{Efficient Adaptation of VLMs}
        Given the high cost of full fine-tuning and the associated risk of catastrophic forgetting, Parameter-Efficient Fine-Tuning (PEFT) \cite{jung2025gralora, zou2025flylora} has emerged as a standard for adapting VLMs to downstream tasks. Although these methods update only a small number of parameters, they introduce trainable low-rank increments during the forward pass, thereby modifying the effective function of the VLM backbone. Recent works have explored adapter-based architectures \cite{wang2025vlaadapter} to bridge the modality gap in VLA design. However, directly leveraging a generalized VLM for robotic manipulation presents challenges. First, VLMs are pre-trained on internet data primarily for semantic tasks such as Visual Question Answering (VQA). Extracting fine-grained, manipulation-relevant features from the high-level representations of VLM is difficult. Second, efficient adaptation methods often introduce insufficient trainable parameters, limiting the model capacity required to learn complex, diverse manipulation trajectories. To address these bottlenecks, we introduce the Mixture of Layers (MoL) to extract manipulation-critical features, and the Meta Skill Module (MSM) to enhance model capacity  with meta skill retrieval and action refinement.

%% file: sections/method.tex
\section{METHOD}
        In this section, we present the $M^2$-VLA. We posit that a general-purpose Vision-Language Model (VLM) can be effectively adapted to serve as a powerful backbone for robotic manipulation. We first establish the problem formulation and introduces the overall $M^2$-VLA architecture. Subsequently, we detail the core strategies: the Mixture of Layers (MoL) and the Meta Skill Module (MSM). The MoL bridges the information gap between semantic representations and detailed actions, while the MSM enhances manipulation capabilities via meta skill retrieval and action refinement.

        \subsection{Preliminary}
        We formulate the robot manipulation task as a conditional decision-making problem. The robot receives a visual observation $\mathcal{O}$, a natural language instruction $\mathcal{L}$ and the proprioceptive state $\mathcal{P}$, where the visual observation $\mathcal{O}$ typically comprises a third-person view $I^{global}$ and a wrist-mounted view $I^{wrist}$. The objective of this problem is to generate a sequence of actions $\mathbf{a_r}$ to complete the specified task. Adopting the Vision-Language-Action (VLA) paradigm, let $\mathcal{E}_{vlm}$ denote a pre-trained VLM backbone, then the whole process can be abstracted as learning a policy $\pi_\theta$:

\begin{align}
\mathbf{a_r} = \pi_\theta(\mathcal{E}_{vlm}(I^{global}, I^{wrist}, \mathcal{L}), \mathcal{P})
\label{eq:policy}
\end{align}

\subsection{Main Structure}
\begin{figure*}[t]
    \centering
    \includegraphics[width=0.70\linewidth]{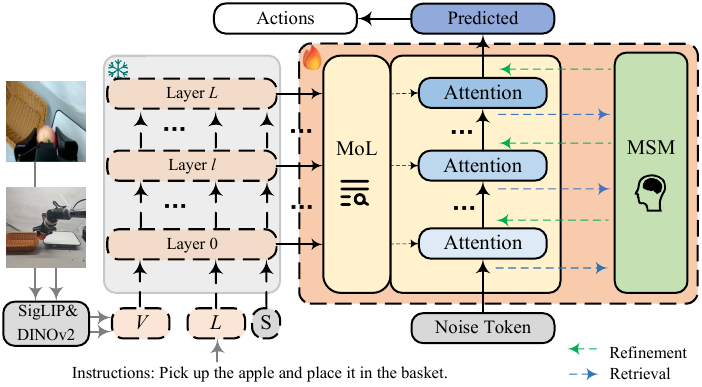}
    \caption{Overview of the $M^2$-VLA framework. The model extracts visual and language features, concatenates them with learnable queries, and utilizes a pre-trained VLM as the perceptual backbone. Action generation is decoupled and performed via a denoising transformer head, with Mixture of Layers (MoL) performing feature extraction from VLM, and Meta Skill Module (MSM) improving model capacity.}
    \label{fig:framework}
\end{figure*}

As discussed in Section~\ref{sec:intro}, the paradigm of fine-tuning VLM on specific robotic action datasets leads to poor generalization. To address this, we employ a pre-trained VLM as robust perceptual backbone, thereby preserving its generalization capabilities. Furthermore, in contrast to approaches that force the VLM to directly autoregress actions, our approach decouples the action generation expert from the perceptual backbone. Overall, $M^2$-VLA consists of two primary streams: \textbf{Feature Extraction} and \textbf{Action Generation}. Fig.~\ref{fig:framework} illustrates the framework of $M^2$-VLA.
 
\noindent\textbf{Feature Extraction}. To capture comprehensive spatial and semantic information, we process the visual observation $I^{global}$ and $I^{wrist}$ using DINOv2 and SigLIP simultaneously, yielding visual embeddings $\mathcal{V}$. Concurrently, the language instruction $\mathcal{L}$ is tokenized into text embeddings $\mathcal{L}_{emb}$.
To explicitly extract visual-textual information from a large VLM \cite{cui2025openhelix,wang2025vlaadapter}, we introduce a set of learnable queries, $\mathcal{S}$, appended to the input sequence to capture cross-modal interactions with both visual and language tokens.
The combined input is fed into the VLM backbone as:
\begin{align}
\mathcal{H} = \mathcal{E}_{vlm}([\mathcal{V}, \mathcal{L}_{emb}, \mathcal{S}])
\label{eq:hlm}
\end{align}
\noindent For a backbone with $L$ layers, we obtain the multi-scale hidden states $\mathcal{H} = \{(\mathcal{V}'_l, \mathcal{L}'_l, \mathcal{S}'_l)\}_{l=1}^{L}$, representing the processed visual, text, and query features at each layer. These features together with the proprioceptive state $\mathcal{P}$ provide the conditioning context for downstream action generation.

\noindent\textbf{Action Generation}. We design the action head as a multi-layer transformer, and model the action generation process as a layer-wise iterative reconstruction process.
During training, we sample gaussian noise $\epsilon \sim \mathcal{N}(0, \mathbf{I})$ to initialize the noisy action $\mathbf{a}_n$. Conditioned on the features extracted from the VLM backbone, the action head is trained to predict the injected noise and reconstruct the target action $\mathbf{a}_r$.

\begin{align}
\mathbf{a_r}= \mathbf{a}_n - \mathcal{D}_\phi(\mathbf{a}_n, \mathcal{H}, \mathcal{P})
\label{eq:denoise}
\end{align}

\noindent where $\mathcal{D}_\phi$ is the action head network with trainable parameters $\phi$. And the action head internally maps action to a latent vector $\mathcal{A}$ through linear transformations.

However, extracting manipulation-relevant information from the VLM's large internal representations remains challenging. First, the VLM encodes abundant semantic and spatial information, a substantial portion of which is redundant or irrelevant to low-level control. Second, the lightweight action head often lacks sufficient capacity to effectively process such diverse high-dimensional features while handling a wide range of manipulation tasks. To bridge the gap between VLM semantics and precise control, we introduce the Mixture of Layers (MoL) in Section~\ref{sec:mol}, which filters and selects task-relevant information from each layer of the VLM. To mitigate the capacity bottleneck of the lightweight action head, we propose the Meta Skill Module (MSM) in Section~\ref{sec:msm}, which retrieves and refer to meta skill from historical data to improve generalization ability.

\noindent\textbf{Model Training}. Under the proposed architecture, the VLM parameters are kept frozen to mitigate catastrophic forgetting. Following the supervised learning paradigm, the training objective is formulated as:

\begin{align}
\mathcal{L} = \mathbb{E}_{\mathcal{H}} [ || \mathbf{a}_g - (\mathbf{a}_n - \mathcal{D}_\phi(\mathbf{a}_n, \mathcal{H}, \mathcal{P}))||_1 ]
\label{eq:loss}
\end{align}
\noindent where $\mathbf{a}_g$ is the ground-truth action. Since the parameters of the backbone remain unchanged, $M^2$-VLA preserves the original VLM's full visual language generalization capability and VQA capability.

\subsection{Feature Selection via Mixture-of-Layers}
\label{sec:mol}
The gap between VLM representations and robotic manipulation introduces substantial task-irrelevant noise. However, the high-dimensional VLM features still contain a subset of information directly relevant to 3D manipulation \cite{cai2025depthlm, fan2025vlm3r}. Filtering these relevant features significantly enhances downstream action generation. Therefore, we extract features from each layer of VLM to guide the generation of the action head.

\begin{figure}[t]
    \centering
    \includegraphics[width=0.6\linewidth]{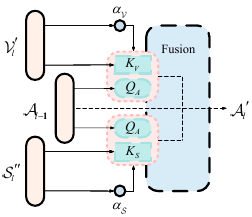}
    \caption{Architecture of the Mixture-of-Layers (MoL). MoL adaptively integrates hierarchical features extracted at various layers of the VLM via three decoupled attention branches, dynamically filtering and aggregating task-relevant knowledge for downstream action generation.}
    \label{fig:mol}
\end{figure}

Different VLM layers encode non-uniform semantic and spatial cues for manipulation. Inspired by the general mixture principle of combining complementary representations, we introduce Mixture of Layers (MoL) to aggregate layer-wise hidden states from the VLM backbone. Illustrated in Fig.~\ref{fig:mol}, the MoL architecture comprises three independent attention mechanisms and its variants.
Specifically, the MoL mechanism is applied at every layer of computation within the action head. For a given layer $l$, let $Q_A$ denote the attention query derived from the previous layer's action latent $\mathcal{A}_{l-1}$. To avoid cross-modal interference during the softmax operation, we decouple the integration process of MoL into three parallel attention branches. In the first branch, we obtain the processed queries $\mathcal{S}'_l$ from the VLM firstly. Concatenated with the proprioceptive state $\mathcal{P}$, $\mathcal{S}''_l = [\mathcal{S}'_l, \mathcal{P}]$ represents compressed visual-textual-proprioceptive information. For computational efficiency, we utilize $\mathcal{S}''_l$ as the key ($K_S$) and value ($V_S$). To dynamically filter information, we compute a head-wise factor $\alpha_{\mathcal{S}}$ conditioned on the $\mathcal{S}''_l$:
\begin{align}
\alpha_{\mathcal{S}} = \text{sigmoid}(W_{{\mathcal{S}}}\mathcal{S}''_l)
\label{eq:alpha}
\end{align}
Subsequently, the action head modulates the attention mechanism via $\alpha_{\mathcal{S}}$ to selectively aggregate features.
\begin{align}
\mathcal{A}_s = \text{softmax}\left(\frac{\alpha_{\mathcal{S}} Q_A K_S}{\sqrt{d_k}}\right)V_S
\label{eq:As}
\end{align}
where $d_k$ is the hidden dimension of the action head.

Meanwhile, the second attention branch aims to incorporate layer-wise visual information $\mathcal{V}'_l$. However, given the high dimensionality of $\mathcal{V}'_l$, computing a dynamic factor would incur significant computational overhead. Therefore, we employ a learnable factor $\alpha_v$ as the filter to regulate the attention scores:
\begin{align}
\mathcal{A}_v = \text{softmax}\left(\frac{\alpha_v Q_A K_V}{\sqrt{d_k}}\right)V_V
\label{eq:Av}
\end{align}
Finally, the third branch performs self-attention on the action latents to suppress noise and obtain $\mathcal{A}_a$. After that, we aggregate the outputs from these three branches to obtain the final representation:
\begin{align}
\mathcal{A}'_l = \text{Mean}(\mathcal{A}_s, \mathcal{A}_v, \mathcal{A}_a)
\label{eq:A-mean}
\end{align}
This decoupled attention design avoids scale inconsistencies between modalities in softmax operation. 

\subsection{Model Capacity Enhancement via Meta Skill Module}
\label{sec:msm}
\begin{figure}[t]
    \centering
    \includegraphics[width=1\linewidth]{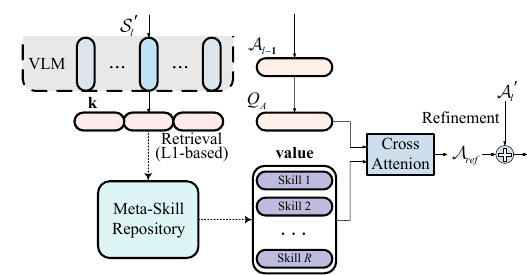}
    \caption{Architecture of the Meta Skill Module (MSM). MSM augments the action head with perceptual-skill pairs, allowing retrieval and integration of relevant meta-skills for action refinement during training and inference.}
    \label{fig:msm}
\end{figure}

The model capacity is to some extent related to the number of parameters, consequently, the lightweight action head design directly may limit its generalization capabilities and overall performance \cite{hoffmann2022training}. To address this bottleneck, we propose the Meta Skill Module (MSM), illustrated in Fig.~\ref{fig:msm}. The MSM operates via two primary mechanisms: \textbf{Memory Construction} during training and \textbf{Skill Retrieval and Action Refinement} during training and inference.

\noindent\textbf{Memory Construction}. We maintain an external queue, denoted as $\mathcal{M}$. Each entry in $\mathcal{M}$ consists of a key-value pair. The key represents the perceptual context, formed by concatenating hierarchical features (shallow, intermediate, and deep layers) from the VLM: 
\begin{align}
\mathbf{k} = [\mathcal{V}'_{low} \oplus \mathcal{V}'_{mid} \oplus \mathcal{V}'_{high}]
\label{eq:key}
\end{align}
The corresponding value is the action sequence that successfully accomplishes the task over the future chunk horizon. In this way, the queue serves as a repository of meta-skills.

\noindent\textbf{Skill Retrieval and Action Refinement}. During the denoising of the action latent, MSM leverages the meta-skills to refine the action latent. Specifically, MSM utilizes the multi-level features of the current observation as a query, denoted as $\mathbf{q}_{obs}$. Then, MSM retrieves the top-R (e.g., $R=4$) nearest neighbors from $\mathcal{M}$ using the $L_1$ distance:

\begin{align}
\mathcal{N}_R = \underset{(\mathbf{k}_{i}, \mathbf{v}_i) \in \mathcal{M}}{\text{Top-R}} \left( -\|\mathbf{q}_{obs} - \mathbf{k}_i\|_1 \right)
\label{eq:topk}
\end{align}

\noindent The retrieved meta-skills $\{\mathbf{value}_i\}_{i=1}^R$ are projected into the latent space to serve as context $\{V^\mathcal{M}_i\}_{i=1}^R, \{K^\mathcal{M}_i\}_{i=1}^R$. Then, MSM employs a standard cross-attention mechanism to obtain the reference item $\mathbf{A}_{ref}$.
\begin{align}
\mathbf{A}_{ref} = \text{softmax}(\frac{Q_AK^\mathcal{M}}{\sqrt{d_k}})V^\mathcal{M}
\label{eq:oref}
\end{align}

\noindent where the current action latent $Q_A$ serves as the query, and the retrieved meta-skills serve as keys and values.
This retrieved reference item is integrated into the primary action flow via a residual learning paradigm, guiding the action reconstruction with proven meta-skills.

\begin{align}
\mathcal{A}_l = \mathcal{A}_{l-1} - (\mathcal{A}'_l + \mathcal{A}_{ref})
\label{eq:A-ref}
\end{align}

\noindent By explicitly referring to mature meta-skills, the MSM reduces the search space for the policy and improves the
policy's ability to learn complex trajectories. In the current implementation, $\mathcal{M}$ is a fixed-size queue, and retrieval costs $O(|\mathcal{M}|d)$ per query. Scaling MSM to lifelong learning with memory pruning or approximate nearest-neighbor retrieval is our future work.

%% file: sections/exp.tex
\section{EXPERIMENTS}
To validate the hypothesis that a generalized VLM can be adapted into a backbone for high-performance robotic manipulation, we conducted experiments below. \textbf{A,} we evaluate the generalization performance of $M^2$-VLA, against existing approaches in settings requiring instruction following and novel objects generalization, showing the necessity of the proposed paradigm. \textbf{B,} we evaluate $M^2$-VLA against existing methods on standard simulation suites, demonstrating its effectiveness. \textbf{C,} we validate the practical applicability and robustness of $M^2$-VLA through real-world experiments. \textbf{D,} we conduct visualization and ablation studies to investigate the interpretability and effectiveness of the proposed strategies. All experiments are performed on 4 NVIDIA A800 GPUs. Thanks to the parameter efficient architecture design, $M^2$-VLA is able to train a powerful VLA model in just 8 hours on a single RTX 3090 GPU.

\subsection{Performance on Generalization Tasks}
The strength of VLM and VLA lies in their ability to generalize and understand language instructions and solve novel object tasks. Therefore, this paper presents our argument and constructs $M^2$-VLA. In this section, we evaluate the generalization performance of $M^2$-VLA from two perspectives: Instruction following and Novel object tasks.

\noindent\textbf{Instruction following}. We compare the proposed method with existing approaches across tasks sharing identical visual observations but significantly differing task instructions. Specifically, each method is trained on the spatial suite of the Libero dataset for 5000 steps and tested on its spatial suite test set. However, we assess the success rate using the synonymically rephrased instructions, and quantify the performance drop relative to the original instructions. Results are recorded in
Table~\ref{tab:spatial_approximate_semantics}.


\begin{table}[H]
\centering
\caption{Comparison of the Instruction Following Capabilities. \textbf{Bold} is the Best Performance.}
\begin{tabular}{l c c}
\toprule
{Method} & \begin{tabular}[c]{@{}c@{}}{Synonymically} \\ {Rephrased Instructions (\%)}\end{tabular} & \begin{tabular}[c]{@{}c@{}}{Performance} \\ {Drop (\%)}\end{tabular} \\
\midrule
OpenVLA \cite{kim2024openvla} & 20.0 & 64.7 \\
VLA-Adapter \cite{wang2025vlaadapter} & 52.8 & 45.4 \\
\rowcolor{gray!15} $M^2$-VLA & \textbf{66.2} & \textbf{29.4} \\
\bottomrule
\end{tabular}
\label{tab:spatial_approximate_semantics}
\end{table}


It can be observed that under conditions of significant language variation, most existing approaches exhibit a sharp decline in performance, whereas the proposed $M^2$-VLA maintains excellent success rate.

\noindent\textbf{Novel Object Tasks}. In manipulation tasks, the objects and visual conditions are varied. In this part, we evaluate the methods' zero-shot ability when solving tasks with novel object and corresponding novel instruction. Specifically, each method was trained on the Libero object suite for 15000 steps, then tested on the novel pick-and-place task. We assess the success rate solving task with novel object: "pick wine bottle...", and quantify the performance drop relative to original tasks. The results are summarized in Table~\ref{tab:unseen_pick_winebottle_eng}.
\begin{table}[H]
\centering
\caption{Comparison on Novel Object Tasks. \textbf{Bold} is the Best Performance}
\begin{tabular}{l c c}
\toprule
{Method} & {Novel Object (\%)} & {Performance Drop (\%)} \\
\midrule
OpenVLA \cite{kim2024openvla}       & 8.2   & 80.2   \\
VLA-Adapter \cite{wang2025vlaadapter}   & 8.0  & 91.6   \\
\rowcolor{gray!15} $M^2$-VLA        & \textbf{34.4}   & \textbf{30.4} \\
\bottomrule
\end{tabular}
\label{tab:unseen_pick_winebottle_eng}
\end{table}

We can observe that, fine-tuning methods typically encounter steep performance drops when evaluated on novel objects. In contrast, $M^2$-VLA preserves its generalization capacity as it employs generalized VLM as backbone. 

\subsection{Performance in Simulation Environments}
\noindent\textbf{Experimental Setup}. To fairly evaluate the effectiveness of $M^2$-VLA, this section compares its task success rates against mature baselines and state-of-the-art methods in simulation environments.We select the Libero simulation environment,  training for 15000 steps and testing for 500 times. All four Libero suites Spatial, Object, Goal and Long are selected for evaluation. For fairness, we primarily select models utilize generalized VLM as the perceptual backbone, that is, using vla-adapter (frozen) \cite{wang2025vlaadapter} and smolVLA \cite {cadene2024smolvla} as baselines. Additionally, state-of-the-art models are selected, including OpenVLA\cite {kim2024openvla}, MolmoAct\cite{molmoact2025}, ThinkAct \cite{thinkact2025}, CoT-VLA \cite{zhao2025cotvla} and FlowVLA \cite{zhong2025flowvla}.

\noindent\textbf{Results}. The results are recorded in Table~\ref{tab:sota_comparison}.
\begin{table*}[ht]
\centering
\caption{Comparison of Simulation Benchmarks Based on Success Rate (\%). "VLM Fine-tuning" Indicates Whether the Method Requires Fine-tuning the VLM Backbone. \textbf{Bold} is the Best Performance.}
\begin{tabular}{lccccccc}
\toprule
{Method} & {VLM Fine-tuning} & {Spatial} & {Object} & {Goal} & {Long} & {Avg.} & {Train. Params (B)} \\
\midrule
OpenVLA \cite{kim2024openvla}             & Yes & 84.7  & 88.4    & 79.2   & 53.7   & 76.5   &   7.0     \\
MolmoAct \cite{molmoact2025}            & Yes & 87.0  & 95.4    & 87.6   & 77.2   & 86.6   &   7.0     \\
ThinkAct \cite{thinkact2025}            & Yes & 88.3  & 91.4    & 87.1   & 70.9   & 84.4   &   7.0     \\
CoT-VLA \cite{zhao2025cotvla}            & Yes & 87.5    & 91.6   & 87.6   & 69.0   & 81.1  &    7.0     \\
FlowVLA \cite{zhong2025flowvla}            & Yes & 93.2    & 95.0   & 91.6   & 72.6   & 88.1  &    7.0     \\
\midrule
SmolVLA \cite{cadene2024smolvla} & No  &   93.0    &    94.0     &    91.0    &    77.0    &   88.8    &  2.2       \\
VLA-Adapter (Frozen) \cite{wang2025vlaadapter} & No &   94.4    &     94.2    &  86.8      & 82.0   &    89.4   &   0.5      \\
\rowcolor{gray!15} $M^2$-VLA              & No  & \textbf{97.8}  & \textbf{99.0}    & \textbf{97.2} & \textbf{87.0} &   \textbf{95.3}   &  \textbf{0.3}       \\
\bottomrule
\end{tabular}
\label{tab:sota_comparison}
\end{table*}
The results demonstrate that $M^2$-VLA achieves significant advantages over other methods that utilize generalized VLM as the backbone. Furthermore, our 0.3B trainable parameter variant achieves performance comparable to 7B-scale SOTA methods. These findings validate the effectiveness of the proposed model.

\subsection{Performance in Real-world}
To further validate the effectiveness of the lightweight $M^2$-VLA on real systems, we conduct real-world manipulation experiments.

\begin{figure*}[ht]
    \centering
    \includegraphics[width=0.8\linewidth]{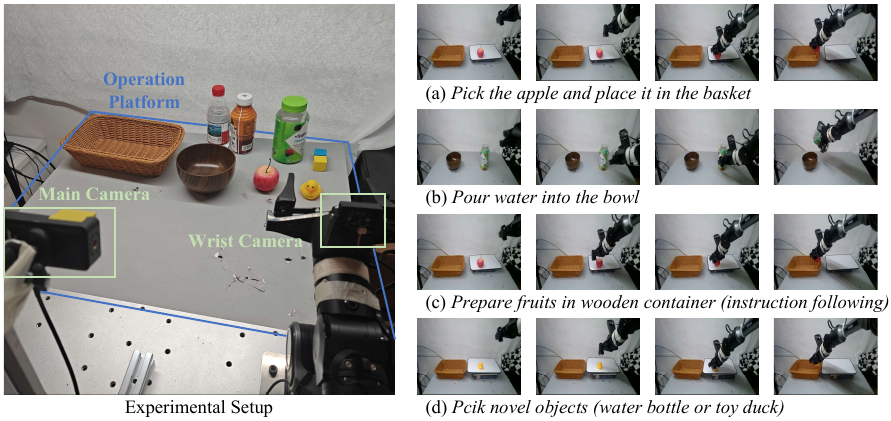}
    \caption{Experimental setup and representative results for real-world manipulation tasks. The sequential images showcase the trajectories of a robotic arm successfully executing four tasks.}
    \label{fig:real_setting}
\end{figure*}

\noindent\textbf{Experimental Setup}. All experiments are deployed on an AgileX PiPer 6-DoF robotic manipulator equipped with a parallel gripper, resulting in a 7-DoF system (illustrated in Fig.~\ref{fig:real_setting}). Visual observations are acquired via 2 RGB cameras: a static front-view camera (main camera) and a wrist-mounted camera (wrist camera).

\noindent\textbf{Evaluation Tasks}. To assess real-world performance and generalization capabilities, inspired by recent works \cite{zhou2024reconvla, liu2024rdt}, we design four categories of evaluation tasks: (1) \textit{pick the apple and place it in the basket}, (2) \textit{pour water into the bowl}, (3) \textit{pick with instruction following}, and (4) \textit{pick novel objects}. (1) and (2) are fundamental tasks, while (3) and (4) are generalization tasks. Specifically, the (3) \textit{pick with instruction following} task evaluates language generalization by providing synonymically rephrased instructions not seen during training. The (4) \textit{pick novel objects} setting evaluates the method's zero-shot ability. Under this setting, the model is trained exclusively on pick-and-place tasks with an apple but is evaluated on novel target objects (a water bottle or a toy duck) accompanied by corresponding novel instructions.

\noindent\textbf{Data Collection and Training}. For each task, we collect 50 demonstration trajectories via teleoperation. For a fair comparison, SmolVLA\cite{cadene2024smolvla}, VLA-Adapter (Frozen) \cite{wang2025vlaadapter}, VLA-Adapter \cite{wang2025vlaadapter} and $M^2$-VLA are trained for 5000 steps.

\noindent\textbf{Results and Analysis}. Each task is evaluated with 20 real-world rollouts. The success rates are reported in Table~\ref{tab:real_world_successes}. Meanwhile, the sequential images in Fig.~\ref{fig:real_setting} showcase the trajectories of a robotic arm successfully executing four tasks.

\begin{table}[h]
\centering
\caption{Comparison of Real-world Tasks Based on Success Rate (\%). \textbf{Bold} is the Best Performance.}
\setlength{\tabcolsep}{2.5pt} 
\begin{tabular}{@{}l@{\hskip 2pt}c@{\hskip 2pt}c|c@{\hskip 2pt}c@{}}
\toprule
& \multicolumn{2}{c|}{{Fundamental}} & \multicolumn{2}{c}{{Generalization}} \\
\cmidrule(lr){2-3}\cmidrule(lr){4-5}
{Method} & 
\begin{tabular}{@{}c@{}}{Pick apple}\\{into basket}\end{tabular} & 
\begin{tabular}{@{}c@{}}{Pour water}\\{into bowl}\end{tabular} & 
\begin{tabular}{@{}c@{}}{Instruction}\\{following}\end{tabular} & 
\begin{tabular}{@{}c@{}}Novel\\ objects\end{tabular} \\
\midrule
smolVLA \cite{cadene2024smolvla} & 25 & 30 & 25 & 20 \\
\begin{tabular}{@{}l@{}}VLA-Adapter\\(Frozen)\end{tabular} \cite{wang2025vlaadapter} & 10 & 0 & 5 & 5 \\
VLA-Adapter \cite{wang2025vlaadapter} & 60 & 50 & 55 & 0 \\
\rowcolor{gray!15} $M^2$-VLA & \textbf{85} & \textbf{90} & \textbf{80} & \textbf{75} \\
\bottomrule
\end{tabular}
\setlength{\tabcolsep}{6pt} 
\label{tab:real_world_successes}
\end{table}
Based on the results, we can draw the following conclusions. Firstly, on standard manipulation tasks, our proposed $M^2$-VLA achieves a high success rate, outperforming all baseline methods. Secondly, under the \textit{pick with instruction following} setting, $M^2$-VLA demonstrates superior robustness to language variations compared to existing approaches. Thirdly, for \textit{novel objects}, while conventional fine-tuning paradigms suffer from performance degradation, methods leveraging generalized VLMs as backbones (i.e., VLA-Adapter (Frozen) and $M^2$-VLA) exhibit stronger generalization without steep performance drops. Furthermore, due to the architectural designs that are specifically customized to adapting generalized VLM backbones, $M^2$-VLA achieves significantly higher success rates. 

Qualitative observations from the \textit{pick novel objects} setting provide further insights. Existing paradigms, such as smolVLA and VLA-Adapter, are easy to overfitting on the training trajectories. Specifically, we observe that the VLA-Adapter frequently exhibits erratic end-gripper flailing when approaching novel targets, ultimately leading to task failure. We believe that this instability stems from a significant distribution shift between the wrist-camera observations and the training data. In contrast, the generalized VLM backbone employed in our approach consistently extracts invariant manipulation features across novel objects, allowing $M^2$-VLA to maintain robust execution and high success rates.

\subsection{Visualization Analysis}
To investigate whether the proposed strategy effectively overcomes bottlenecks encountered during adapting VLM backbone, this section conducts 2 visualization analysis. First,  we verify whether MoL successfully identifies layer experts proficient in spatial information from the massive VLM data. Then, we discusses whether the skill retrieval in MSM can effectively select the approximate meta-skills.

\begin{figure}[htbp]
    \centering
    \includegraphics[width=1\linewidth]{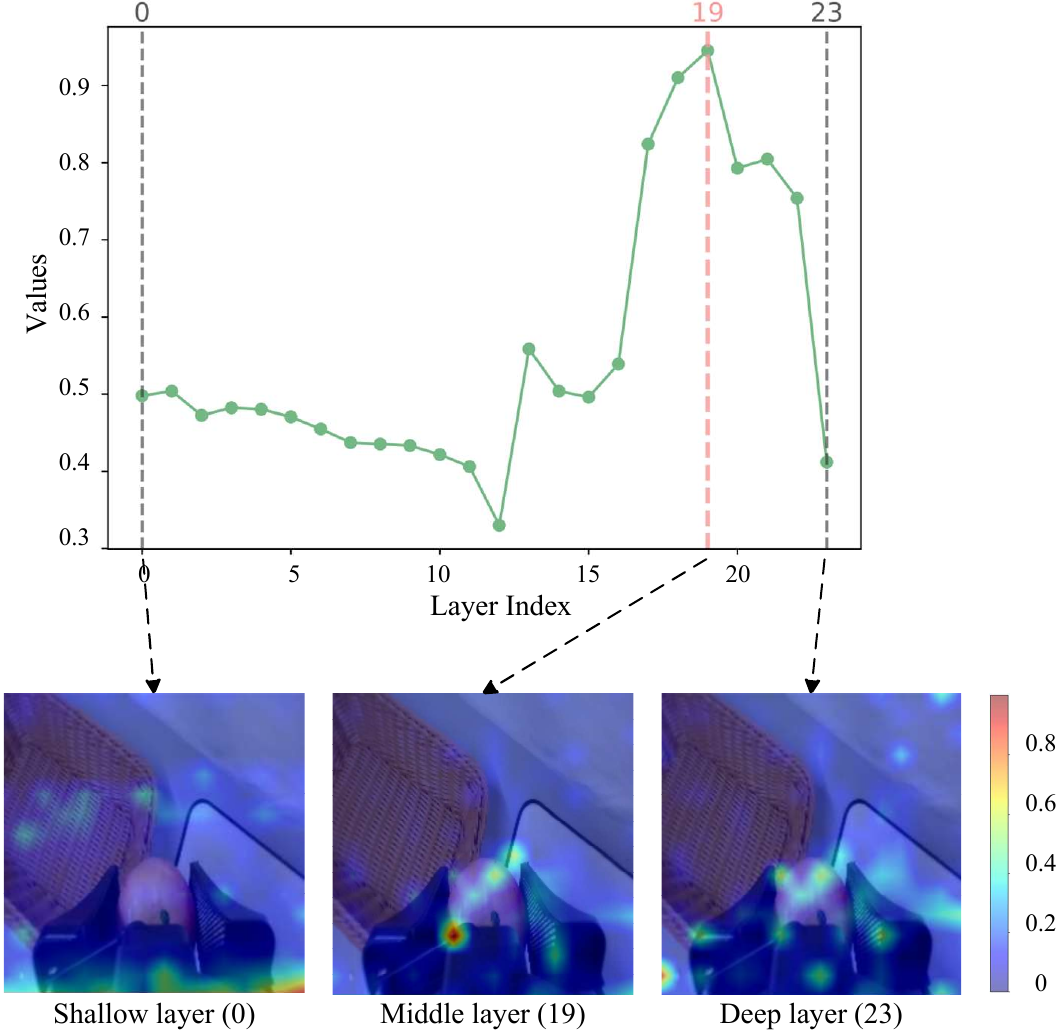}
    \caption{Visualization of the attention heatmap of instruction at shallow layer 0, middle layer 19 and deep layer 23.}
    \label{fig:pick_attention}
\end{figure}

Based on the trained $M^2$-VLA, when performing the \textit{pick the apple and place it in the basket} task, the corresponding $\alpha_{\mathcal{S}}$ values for each layer are plotted in the upper part of  Fig.~\ref{fig:pick_attention}.


In shallow layer 0, middle layer 19, and deep layer 23, the attention heatmap of instruction over observed images are visualized as the lower part of Fig.~\ref{fig:pick_attention}.

We can observe that, the MoL gives more attention on the middle layers. Meanwhile, the brightness of middle layer's attention heatmap is concentrated on the gripper and the target. Although the VLM cannot directly provide the visual information required for robotic operations. However, the MoL successfully identifies operation-relevant visual information from vast information. 

\begin{figure}[htbp]
    \centering
    \includegraphics[width=1\linewidth]{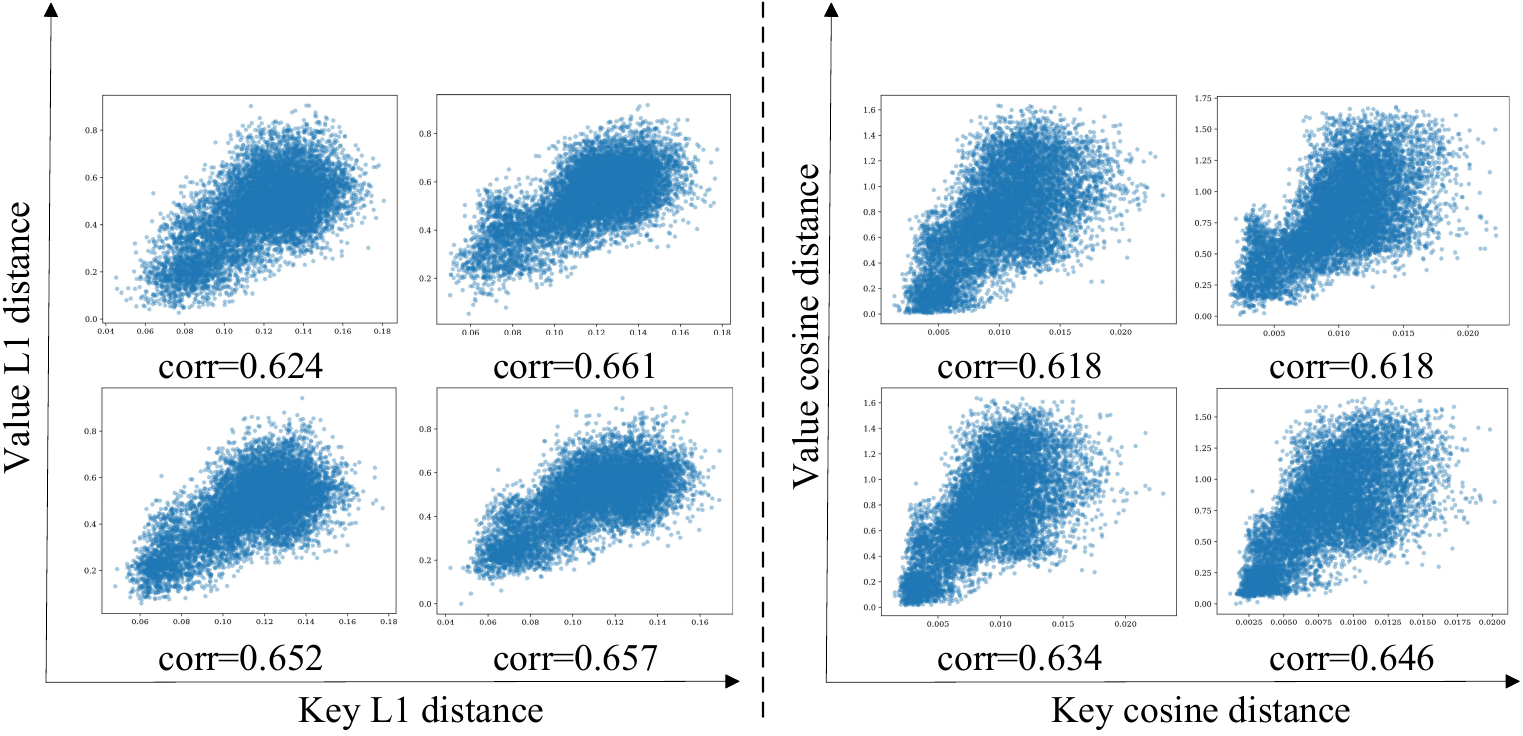}
    \caption{The correlation between Keys and Values within the MSM.}
    \label{fig:msm_distance}
\end{figure}

Next, we visualize the L1 distances between four specific Keys and other Keys within the MSM during model inference, alongside the L1 distances between Values. The cosine distances are also visualized, as shown in Fig.~\ref{fig:msm_distance}. The correlation coefficient (corr) is recorded under each visualization result.

It can be observed that the distances between Keys and between Values exhibit a significant positive correlation, the correlation coefficients are 0.624 and 0.618 respectively. This indicates that the Key construction strategy successfully constructs Keys that retrieval useful skills. Additionally, L1 distance is more effective than cosine distance in distinguishing differences among meta-skills.   
\subsection{Ablation Studies}
\label{sec:ablation}
To further valid the significance of each proposed strategy in overcoming bottlenecks during adapting VLM. We conduct ablation studies of the proposed MoL and MSM on Libero spatial suite in Table~\ref{tab:ablation}, in which the "w/o" means without the corresponding module. All models are trained for 5000 steps.

\begin{table}[h]
\centering
\caption{Ablation Study of MoL and MSM.}
\begin{tabular}{lccc}
\toprule
\textbf{Method} & \textbf{MoL} & \textbf{MSM} & \textbf{success rate (\%)} \\
\midrule
$M^2$-VLA (w/o MoL, w/o MSM) & -- & -- & 46.6 \\
$M^2$-VLA (w/o MSM) & \checkmark & -- & 67.4 \\
$M^2$-VLA (w/o MoL) & -- & \checkmark & 60.2 \\
$M^2$-VLA (full) & \checkmark & \checkmark & 75.0 \\
\bottomrule
\end{tabular}
\label{tab:ablation}
\end{table}

In ablation, $M^2$-VLA performs poorly without MoL and MSM. The MoL assists the Action head in filtering beneficial information for manipulation from the massive information from VLM, achieving significant performance improvements. The MSM enhances the model's capacity with limited parameters, enabling it to handle diverse scenarios.